\documentclass{article}
\usepackage{iclr2027_conference,times}

\usepackage{amsmath,amsfonts,bm}

\def\eqref#1{equation~\ref{#1}}

\def\1{\bm{1}}

\DeclareMathAlphabet{\mathsfit}{\encodingdefault}{\sfdefault}{m}{sl}
\SetMathAlphabet{\mathsfit}{bold}{\encodingdefault}{\sfdefault}{bx}{n}

\usepackage[utf8]{inputenc} 
\usepackage[T1]{fontenc}    
\usepackage{hyperref}       
\usepackage{url}            
\usepackage{booktabs}       
\usepackage{amsfonts}       
\usepackage{nicefrac}       
\usepackage{microtype}      
\usepackage{xcolor}         
\usepackage{graphicx}
\usepackage{amsmath}
\usepackage{subcaption}
\usepackage{placeins}
\usepackage{wrapfig}
\usepackage{float}

\title{Modeling the object representations underlying human physical reasoning}

\author{%
\begin{minipage}[t]{\dimexpr\textwidth-2\tabcolsep\relax}
\centering
\normalfont
\textbf{%
  Andrey Gizdov\textsuperscript{1,2,3}\thanks{Equal contribution.}\quad
  Andrea Procopio\textsuperscript{2,3,4}\footnotemark[1]\quad
  Lorenzo Caputi\textsuperscript{3,5}\footnotemark[1]%
}\\[0.6em]
\textbf{%
  Georgi I. Ivanov\textsuperscript{7}\quad
  Yichen Li\textsuperscript{2}\quad
  Daniel Harari\textsuperscript{4,6}\quad
  Tomer Ullman\textsuperscript{2}%
}\\[1em]
\textsuperscript{1}OpenVector \quad
\textsuperscript{2}Harvard University \quad
\textsuperscript{3}BAINSA \quad
\textsuperscript{4}MIT \\[0.2em]
\textsuperscript{5}Stanford University \quad
\textsuperscript{6}Weizmann Institute of Science \quad
\textsuperscript{7}Theopha
\end{minipage}%
}

\iclrfinalcopy 

\begin{document}

\maketitle
\lhead{}

\begin{abstract}
Humans appear to represent objects when reasoning about physics with coarse, volumetric "bodies" that smooth concavities, trading fine visual detail for efficient physical predictions. Yet, the structure of these representations remains largely unknown. Segmentation models, in contrast, are trained for pixel-accurate masks that may misalign with such bodies. We ask whether and when these models nonetheless acquire human-like object representations. Using a time-to-collision (TTC) and change detection (CD) behavioral task with data from 178 and 50 human participants, respectively, we introduce a pipeline and an alignment metric to compare the visual representations of segmentation models to those of humans. We do this systematically on multiple architectures (DINOv2, SegFormer, DeepLabV3+, and UPerNet), varying their size and training time. We find that briefly trained models segment objects too coarsely, aligning poorly with humans, while fully trained models segment objects too finely. For each model, there is an intermediate training regime that best matches the coarse bodies observed in human behaviour, and larger models tend to reach it earlier. We show these bodies emerge under resource constraints in general-purpose vision models, providing computational support to resource-rational accounts of human cognition. This work provides a foundational framework for testing alignment between vision models and humans and shows there is a growing gap between the state-of-the-art in artificial intelligence and human cognition, driven by scaling model size and training.

\end{abstract}

\section{Introduction}

Human perception is concerned with \emph{what} entities are present, \emph{where} the entities are, and \emph{how} a scene will unfold ~\citep{marr2010vision, freyd1987dynamic}. The problems of `what', `where', and `how' are coupled but separate, and it is likely that they are supported by  computations and representations that are also coupled but separate. In particular, for the purposes of inferring the \textit{identity} of objects (e.g. telling apart a thermos and a water bottle) it may be important to have fine-grain object segmentation, but for the purposes of estimating \textit{how} a physical scene will unfold (which objects will collide, and where they will end up), it may be sufficient and more cost-effective to have more coarse object-representations (e.g. if all that matters is catching or ducking them up, a water bottle and thermos can be roughly approximated by a cylinder).

\begin{figure}[t]
    \centering
    \begin{minipage}[c]{0.62\textwidth}
        \centering
        \includegraphics[width=\linewidth]{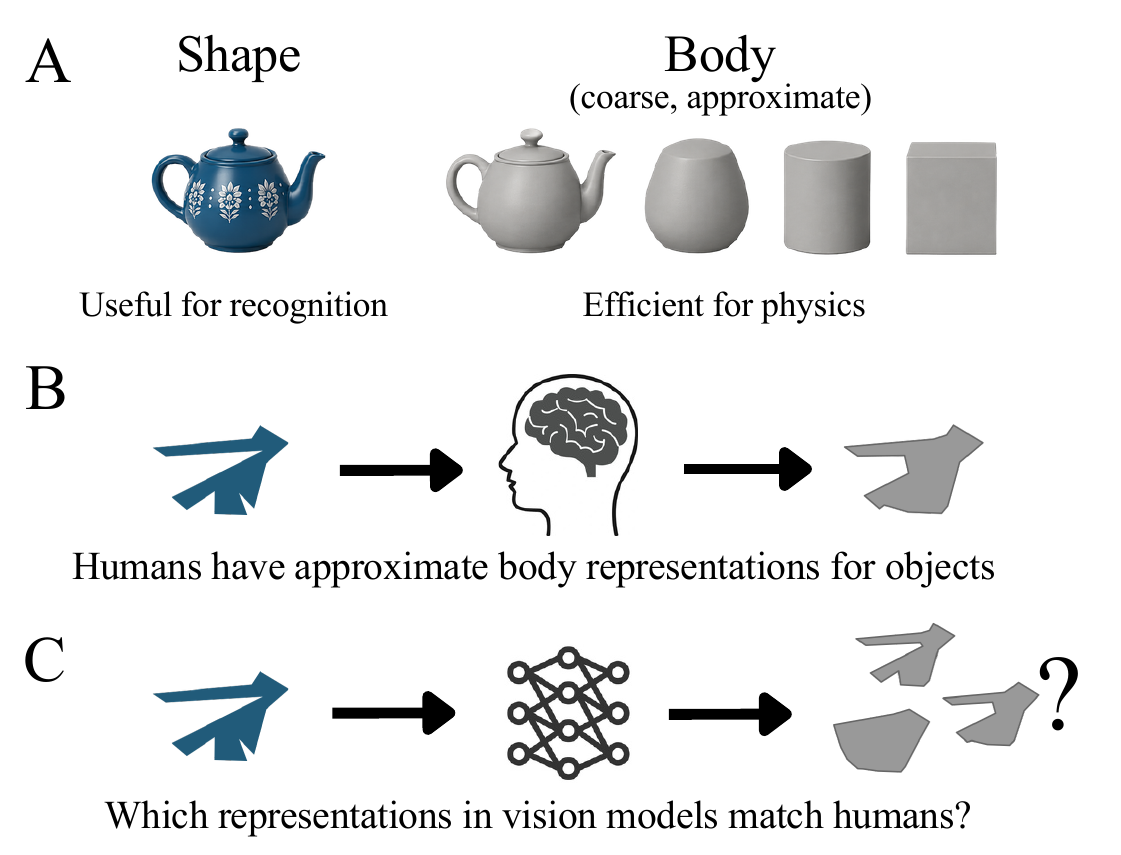}
    \end{minipage}\hfill
    \begin{minipage}[c]{0.35\textwidth}
        \caption{\textbf{Research overview.} \textbf{(A)} Different object representations are useful for different goals. ``Body'' representation is useful for physical reasoning. ``Shape'' representation is useful for recognition. \textbf{(B)} Prior work ~\citep{li2023approximate} has shown evidence that humans do have such coarse body representations for physical reasoning (e.g. predicting collision times), specifically by ``filling in'' concave regions. \textbf{(C)} Our research asks \textit{whether} vision segmentation models have such similar representations and under \textit{what} conditions they develop them.}
        \label{fig:overview}
    \end{minipage}
\end{figure}

The separation between coarse and fine-grain object representations for different purposes is widely used in engineering, especially in simulated environments such as games and animations. Computational models of intuitive physics in humans suggest people may have a similar split between physics and graphics ~\citep{ullman2017mind, balaban2025physics}, with fine-grain object \textit{shapes} used for rendering and recognition and coarse-grain \textit{body} representations used for physical prediction and action (see Figure \ref{fig:overview}A). This suggestion is in line with studies that show infants first carve the visual field into cohesive but rough volumetric entities with approximate spatial extent, before they infer contact relations, support, or motion trajectories~\citep{spelke1990principles, baillargeon2004infant}. Recent psychophysical studies with adults also suggest that humans use coarse body approximations ~\citep{li2023approximate} (see Figure \ref{fig:overview}B). In particular, there's evidence that people's body approximations tend to "fill in" concave object regions (Section~\ref{sec:related_work} discusses this in detail). From a neuro-scientific perspective, the physics/graphics split may map onto the neural division between the dorsal and ventral streams in human vision ~\citep{balaban2025physics}.

While evidence suggests humans rely on approximate representations of objects when reasoning about physics, the nature and structure of these representations in the human brain remain largely unknown. Behavioral methods offer indirect, low-resolution glimpses into these internal encodings, and do not reveal their geometric or computational form. Artificial segmentation models, which are trained to perform a similar decomposition of scenes into discrete entities, offer a computational setting for exploring how representations with similar behavioral consequences might form. If modern vision models trained for segmentation or prediction share representational structure with humans, they could provide a scalable way of testing candidate object approximations against human physical judgments.

\textit{Unlike} the study of alignment in LLMs, there has been relatively little exploration of the alignment between the body representations humans use when reasoning about physics and the object representations  models use for vision, which are often used as a basis for downstream physics tasks in robotics and embodied agents, where the potential benefits of such alignment can be important. Given that (i) coarse object bodies are useful for physical simulations in engineered systems,  
(ii) humans appear to rely on coarse bodies during physical reasoning, and  
(iii) segmentation networks can learn such abstractions,  we ask the following question (Figure \ref{fig:overview}C):  \emph{Do segmentation models form object representations whose behavioral consequences match human physical judgments, and when?}

\textbf{Contributions.} In this study, we create a pipeline for such comparison and explicitly control three key factors: model architecture, size, and training time.
Our contributions are as follows:
\begin{enumerate}
    \item \textbf{A general pipeline for comparing segmentation models with human physical judgments.}  
    We operationalize the outputs of segmentation networks and introduce a unified Time-To-Collision (TTC) and Change Detection (CD) evaluation framework using data from 178 and 50 human participants respectively~\citep{li2023approximate}. This framework enables direct, quantitative comparison between vision model predictions and human behavior in physical reasoning tasks (see Section~\ref{sec:methods}).

    \item \textbf{We find an intermediate training regime whose timing varies with model size.}  
    Across four architecture families (16 model configurations), error with respect to human judgments is non-monotonic over training: briefly trained models produce overly coarse masks, while continued training resolves increasingly fine boundary detail. Held-out participant analysis confirms that error is minimized between these endpoints, and different model sizes reach this regime at different times (see Section~\ref{sec:human_like_representations}).

    \item \textbf{We show that segmentation models develop coarse object approximations whose behavioral consequences resemble human concavity-filling effects.}

    Across models, agreement with human judgments is greatest at intermediate training stages, whose timing varies with model size. Fitted geometric operators applied to the ground-truth masks reproduce the human effect equally well (Section~\ref{sec:geometric_baselines}); the contribution of the models is therefore not a closer fit but an account of how such geometry arises in a general-purpose learner. This motivates the hypothesis that resource constraints may contribute to similar coarse encodings in biology (see Section~\ref{sec:discussion}). We characterize the geometry of these model representations and when they arise during training (see Section~\ref{sec:human_like_representations_are_lim_networks}).
\end{enumerate}

Our results provide practical and theoretical insight. Practically, understanding when models form object approximations that align with human judgments may inform future work on vision systems that interact safely and predictably with humans, particularly as we start deploying robots and embodied agents in the physical world. Theoretically, we identify computational conditions under which human-like body representations emerge, offering a computational and visual account of how vision models can produce geometry consistent with human physical judgments.

\section{Related Work}
\label{sec:related_work}
People reason efficiently about everyday physical dynamics, though they show systematic biases under certain conditions ~\citep{kubricht2017intuitive, hartshorne2025insights}. This ``intuitive physics'' emerges early ~\citep{spelke2007core,baillargeon2014infants}, is supported by dedicated neural systems ~\citep{fischer2016functional, pramod2022invariant, pramod2025decoding,balaban2024electrophysiology}, and is likely shared with non-human animals ~\citep{wood2024object, spelke2022babies}. 

There are different competing formal accounts of the mental computations that support intuitive physics. One currently prominent view suggests that people mentally simulate physical events, somewhat like the game engines or physics engine used to simulate fictional worlds in computer games and animation ~\citep{hegarty2004mechanical, battaglia2013simulation, ullman2017mind}. The mental-game-engine proposal has been used to investigate and explain people's reasoning about a variety of physical phenomena, including collisions, stability, tool use, liquids, rigid- and soft-body motion, physical prediction, and physical counterfactual and causal reasoning ~\citep{smith2024probabilistic, smith2013sources, bramley2018intuitive, sanborn2013reconciling, hamrick2016inferring, bates2019modeling, gerstenberg2021counterfactual, sosa2021moral, battaglia2013simulation}.

Prior to, and proceeding alongside the mental simulation account, a variety of non-simulation accounts of intuitive physics have also been proposed. These include, but are not limited to, the use of heuristics, logical rules, bottom-up visual features, qualitative reasoning, and pre-Newtonian theory-like abstractions ~\citep{lerer2016learning,baillargeon2002acquisition,fragkiadaki2015learning,mottaghi2016newtonian,proffitt1990understanding, nusseck2007perception, gilden1989understanding,siegler1998developmental, davis2014representations, forbus1988qualitative,piloto2022intuitive,mccloskey1983naive}. Some of these accounts point to people's errors as evidence that humans cannot be running a perfect simulation, as such a simulation would be too costly and not make errors. However, if such a mental simulation exists in the mind, it cannot be perfectly accurate. Just as engineered simulations make heavy use of various approximations and workarounds, it is likely that people's mental simulation uses approximations as well ~\citep{ullman2017mind, balaban2025physics, wang2025resource, bass2021partial}. Recent evidence further suggests that simulation and abstraction need not be competing accounts: people may blend them depending on the state and task ~\citep{sosa2025blending}.

One major approximation in simulated environments is the use of simplified objects for the purposes of physical tracking and collision handling. While an advanced game engine may use fine-grain meshes to graphically render high-resolution images, it will often use only rough bodies for the purposes of collision detection and physics simulation. Converging theoretical and empirical work suggests that humans similarly use approximate ``body'' representations in physical reasoning ~\citep{ullman2017mind, li2023approximate} -- volumetric approximations that prioritize properties relevant to action and physical interaction (e.g., mass, position, and rough form) while discarding fine-grained visual details (Figure \ref{fig:overview} A, B). This idea is consistent with resource-rational accounts of cognition~\citep{lieder2020resource}, and the ``what/where–how'' functional split in visual processing ~\citep{ungerleider1982two, goodale1992separate}.

In this paper, we focus on the Time-To-Collision and Change Detection experiments of \citet[Experiments~2 and~3b]{li2023approximate} (Figure~\ref{fig:experiments_given_to_humans}A). The first builds on a well-established paradigm for time-to-collision judgments with simple bodies~\citep{rosenbaum1975perception,tresilian1995perceptual,gray2001exploring}. A key behavioral signature is an asymmetry between \textit{concave} and \textit{convex} stimuli: different from what a random noise model would predict, human judgments seem to selectively smooth or ``fill in'' concavities on objects (over-approximating inlets and gaps), predicting earlier collision for concave than convex pairings. Similarly, in the Change Detection experiment, participants are significantly worse at noticing local geometric additions when they occur inside an object's concavities compared to matched changes on convex protrusions, further supporting the idea that the human visual system mentally smooths over these internal gaps. Preliminary analyses suggested that the operative body representation is between the exact outline and the convex hull.

Segmentation networks provide a controlled computational probe of object form because their outputs expose explicit spatial masks. Yet most of these models are trained for pixel-accurate boundaries, favoring fine detail that may be misaligned with human physical prediction. Prior work has leveraged coarse bodies for modeling human expectations without direct comparison to people ~\citep{smith2019modeling}, or has used $\alpha$-shape–based models to analyze human approximations without endorsing them as cognitive mechanisms ~\citep{li2023approximate}, but as a way of teasing apart different degrees of approximation.

Our work also relates to studies of how alignment with humans changes with model scale and training. In vision, more accurate object-recognition networks use visual strategies less aligned with those of humans~\citep{fel2022harmonizing} and predict neural responses in inferotemporal cortex less well~\citep{linsley2023performance}. In language, larger models fit human reading times worse~\citep{oh2023why}, and fit is best after about two billion training tokens~\citep{oh2023transformer}. Neural networks are also known to learn coarse, low-frequency structure before fine detail~\citep{rahaman2019spectral}. We study what this means for object shape: which geometry segmentation models pass through as they grow or train longer, and when it matches human physical judgments.

\section{Methods}
\label{sec:methods}

Here we outline how we compare vision models with human behavior in two psychophysical tasks: a time-to-collision (TTC) task and a change-detection (CD) task.

\begin{figure}[t]
    \begin{wrapfigure}{l}{0.55\textwidth}
        \vspace{-12pt}
        \centering
        \includegraphics[width=\linewidth]{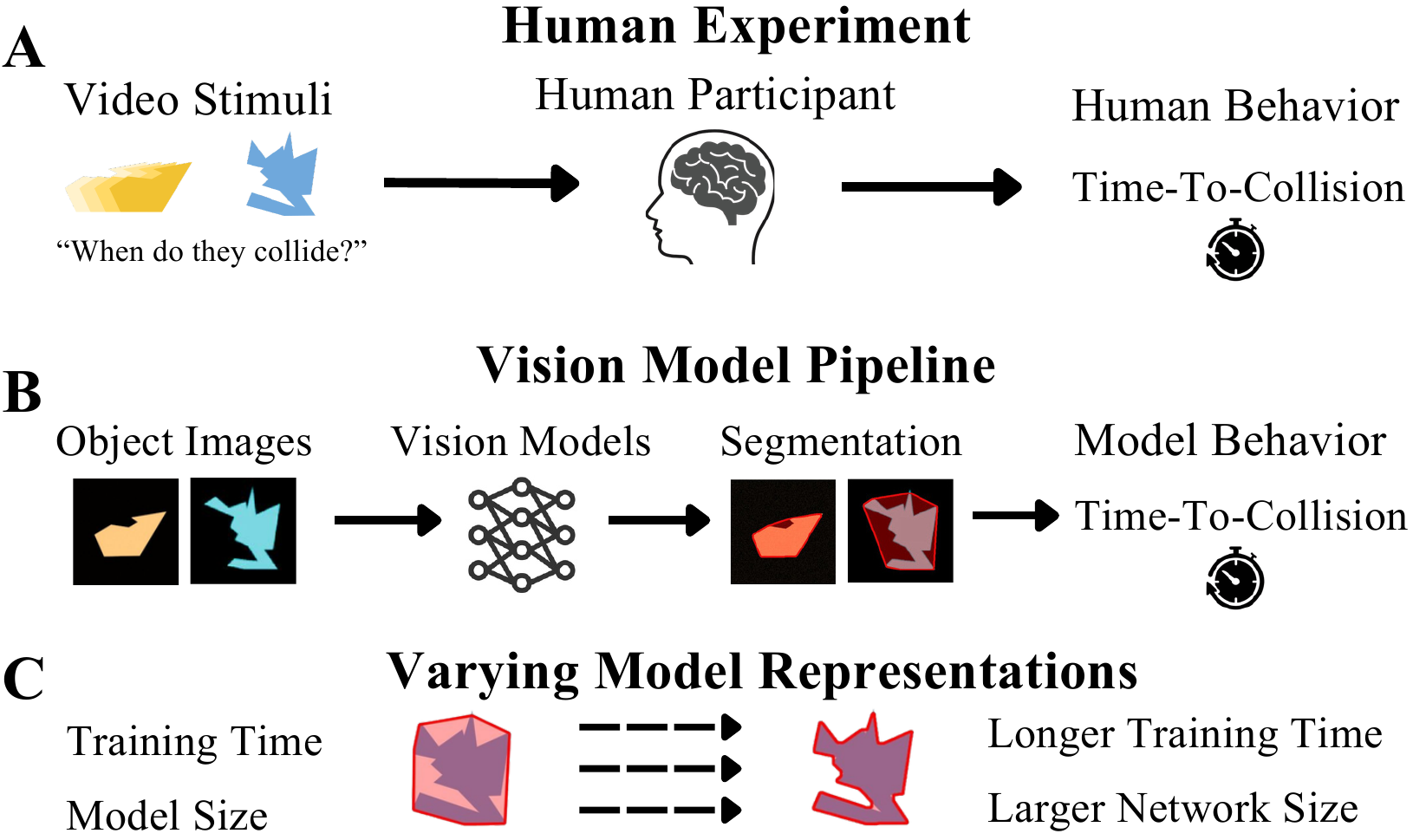}
        \vspace{-14pt}
    \end{wrapfigure}
    \noindent\refstepcounter{figure}\label{fig:experiments_given_to_humans}%
            Figure~\thefigure: \textbf{Methods overview.} \textbf{(A)} In \citet{li2023approximate}, participants predicted when a moving polygon would hit a stationary one that was invisible during motion. At the same true collision time, they responded earlier to hits into a concavity than onto a convex side, as if concavities were filled in. \textbf{(B)} Fine-tuned models segment the frame before motion; we move the masks at the true velocities and record when they overlap. \textbf{(C)} Training time and model size change how coarse the masks are.\par
\end{figure}

\textbf{Human data.} We use the publicly available data of \citet{li2023approximate} from two experiments.
In the \emph{TTC experiment}, 178 participants watched 96 test videos in which one polygon moves toward a stationary one. At a known time the stationary polygon disappears, and participants press a key when they think the two have collided (Figure~\ref{fig:experiments_given_to_humans}A). The videos form 48 matched pairs (8 shapes $\times$ 6 ground-truth collision times). The two videos of a pair share the same kinematics and differ only in whether a concave or a convex side of the stationary polygon faces the collision. Humans respond earlier in the concave condition, consistent with ``filling in'' concavities.
In the \emph{CD experiment}, 50 participants saw a polygon $I_{\mathrm{init}}$ for 1\,s, a blank screen for 2\,s, and a second polygon $I_{\mathrm{out}}$ for 1\,s, and reported whether it had changed. $I_{\mathrm{out}}$ adds a small triangle of equal area either inside a concavity or on a convex part (Appendix~\ref{app:cd_stimuli}). Humans detect convex changes more often than concave ones.
We ask whether segmentation models reproduce both effects, and at which model sizes and training stages (Figure~\ref{fig:experiments_given_to_humans}C).

\textbf{Architectures.} We evaluate four segmentation families (16 configurations): \textbf{SegFormer}~\citep{xie2021segformer} B0--B5 (3.7--84.6M parameters), \textbf{DeepLabV3+}~\citep{chen2018encoder} with ResNet ~\citep{he2016deep} 18/50/101 backbones (12.3--60.4M), \textbf{UPerNet}~\citep{xiao2018unified} with ConvNeXt ~\citep{liu2022convnet} Small/Base/Large/XLarge backbones (82.0--391.0M), and \textbf{DINOv2}~\citep{oquab2024dinov2} Small/Base/Large (27.2--310.6M) with a light convolutional decoder. SegFormer and UPerNet are initialized from checkpoints trained for semantic segmentation on ADE20K~\citep{zhou2017scene}, DeepLabV3+ from Cityscapes~\citep{cordts2016cityscapes}, and DINOv2 from its self-supervised weights. These families cover convolutional (DeepLabV3+, UPerNet) and transformer (SegFormer, DINOv2) backbones, supervised and self-supervised pretraining, and each comes in several sizes, which lets us vary model size within an architecture. All models are fine-tuned end to end on our dataset, with three seeds per configuration.

\textbf{Fine-tuning.} The polygons of \citet{li2023approximate} are irregular shapes chosen for their low verbal association ~\citep{cooper1975mental}, which limits the influence of familiarity with specific objects. The same holds for our segmentation models, which are pretrained on natural images and do not reliably segment these novel polygonal stimuli out of the box. We therefore fine-tune each model on a synthetic polygon dataset that matches the geometric and color statistics of the human stimuli but shares no shapes with them (Appendix~\ref{app:synthetic-dataset}). All models are trained for 945 steps with the same recipe (Appendix~\ref{app:finetuning_hyperparameters}): AdamW ~\citep{loshchilov2019decoupled} with learning rate $5\times10^{-5}$and a cross-entropy plus Dice loss ~\citep{milletari2016vnet}. The human stimuli are used only for evaluation. We release the dataset and its generator with our code.

\subsection{Time-To-Collision}
\label{sec:ttc_pipeline}

Once trained, we test the models on the same videos shown to humans (Figure~\ref{fig:experiments_given_to_humans}B). Before motion starts, both objects are stationary and fully visible. We pass this initial frame $I \in \mathbb{R}^{H \times W \times 3}$ through the model, which outputs per-pixel logits $\mathbf{L}(I) \in \mathbb{R}^{H \times W \times 2}$, and take the per-pixel argmax to obtain a foreground mask:
\begin{equation}
\mathbf{M}(I) = \mathbf{1}\!\left[\operatorname*{arg\,max}_{c \in \{0,1\}} \mathbf{L}(I)_c = 1\right] \in \{0,1\}^{H \times W}.
\end{equation}
Given that each image contains two objects, we extract the two largest connected components of $\mathbf{M}(I)$ and denote them $\mathbf{M}_1, \mathbf{M}_2 \in \{0,1\}^{H \times W}$. Those are the two polygons.

\textbf{From segmentation to video kinematics.} Our goal is to isolate the role of object shape in TTC prediction. We therefore provide the model with the true object velocities $\bm{v}_1, \bm{v}_2 \in \mathbb{R}^2$ (pixels per frame) and the frame rate $\Delta f$ (frames per second) from the stimulus metadata, rather than asking it to estimate motion. $\bm{v}_2 = 0$ naturally, since one object is stationary, but we define it for notation consistency. This makes the model's TTC prediction depend only on its segmentation of the initial frame, so differences in model TTC are attributable to the geometry of the predicted masks, rather than to motion estimation.

We define a translation operator $\operatorname{Translate}(\mathbf{M}, \bm{u})$ that shifts a binary mask $\mathbf{M}$ by displacement $\bm{u}$. We simulate the future motion of the two segmented objects using only their initial masks and the true velocities: at discrete frame index $n \in \mathbb{N}$, $\mathbf{M}_i^{(n)} = \operatorname{Translate}(\mathbf{M}_i, n\bm{v}_i)$ for $i \in \{1, 2\}$. The model-predicted time-to-collision is the earliest time $t = n/\Delta f$ at which the translated masks first overlap:
\begin{equation}
\mathrm{TTC}_{\mathrm{model}} = \min_{n \ge 0} \left\{\, t = \tfrac{n}{\Delta f} \;\middle|\; \mathbf{M}_1^{(n)} \cap \mathbf{M}_2^{(n)} \neq \emptyset \,\right\}.
\end{equation}
The model thus receives the same static pre-motion frame as the human observer and uses only its inferred object masks, together with the true kinematics, to produce a TTC prediction.

\textbf{Comparison to human behavior.} The human dataset contains matched video pairs that share the same kinematics and ground-truth collision time $\mathrm{TTC}_{\mathrm{gt}}$, and differ only in whether a concave or a convex side of one object faces the collision. For each video $j$ and participant $p$, we observe a human response time $\mathrm{TTC}_{\mathrm{human}}^{(j,p)}$, defined as the time between motion onset and button press, and obtain a per-video human TTC by averaging across participants: $\mathrm{TTC}_{\mathrm{human}}^{(j)} = \tfrac{1}{P}\sum_{p=1}^{P} \mathrm{TTC}_{\mathrm{human}}^{(j,p)}$. Indexing videos by ground-truth collision time $\tau$ and shape condition $g \in \{\mathrm{concave}, \mathrm{convex}\}$, and letting $\mathcal{N}(\tau, g)$ denote the corresponding video set, we define the condition-averaged TTC for humans and models:
\begin{equation}
\mathrm{TTC}_x(\tau, g) = \frac{1}{|\mathcal{N}(\tau, g)|} \sum_{j \in \mathcal{N}(\tau, g)} \mathrm{TTC}_x^{(j)}, \qquad x \in \{\mathrm{human}, \mathrm{model}\}.
\end{equation}
The \emph{concavity effect} $\Delta_x(\tau)$ is the difference in TTC between the convex and concave conditions at a given ground-truth time $\tau$:
\begin{equation}
\Delta_x(\tau) = \mathrm{TTC}_x(\tau, \mathrm{convex}) - \mathrm{TTC}_x(\tau, \mathrm{concave}).
\label{eq:delta}
\end{equation}

A positive concavity effect indicates earlier predicted or reported
collisions for concave than convex configurations; a negative effect
indicates the reverse. For each training seed $s$ at a given checkpoint, we define the signed
discrepancy from the human concavity effect as
\begin{equation}
E_s =
\frac{1}{|\mathcal{T}|}
\sum_{\tau \in \mathcal{T}}
\left[
\Delta_{\mathrm{model},s}(\tau)
-
\Delta_{\mathrm{human}}(\tau)
\right],
\label{eq:ttc_signed_error}
\end{equation}
where $\mathcal{T}$ is the set of six ground-truth collision times and $s$ indexes independently trained models (seeds).
$\Delta_{\mathrm{human}}(\tau)$ is computed from all 178 participants, except in the cross-validation of Section~\ref{sec:human_like_representations}, where it is computed only from the participants in the relevant fold. Averaged over participants, the human effect is $31.17$\,ms (95\% CI $[22.74, 39.60]$, $t$-interval over the 178 per-participant convex$-$concave differences).
Positive $E_s$ means a larger mean concavity effect than humans, negative $E_s$ a smaller one; agreement occurs at $E_s = 0$. We therefore use $|E_s|$ as the alignment error of seed $s$, and report its mean over seeds, $\bar{E}$ (both in ms). Because the absolute value is taken after averaging over collision times, a low $\bar{E}$ means agreement in the \emph{mean} concavity effect, not at every collision time, and does not imply that model and human representations are identical. Section~\ref{sec:human_like_representations} reports $\bar{E}$ across training, with the best checkpoint selected on a random half of the participants and evaluated on the other half (200 splits, both directions); Appendix~\ref{app:ttc_concavity_effect} reports the signed effect $\Delta$.

\subsection{Change Detection}
\label{sec:cd_pipeline}

Each trial provides two images: an initial polygon $I_{\mathrm{init}}$ and a second polygon $I_{\mathrm{out}}$ that differs from it by a small segment added to the boundary, either inside a concavity or on a convex protrusion (16 stimuli per condition; examples in Appendix~\ref{app:cd_stimuli}). We pass each image through the segmentation model as in Section~\ref{sec:ttc_pipeline}, and denote by $A_{\mathrm{init}}$ and $A_{\mathrm{out}}$ the areas of the two predicted masks. Whereas the TTC task measures shape representation indirectly through simplified physical simulation, this task measures it directly through the model's mask sensitivity to local shape changes.

\textbf{Relative Area Change (RAC).} Let $A_{\mathrm{gt}}$ be the ground-truth pixel area of the added segment. We define the Relative Area Change as:
\begin{equation}
\mathrm{RAC} = \frac{A_{\mathrm{out}} - A_{\mathrm{init}}}{A_{\mathrm{gt}}}.
\end{equation}
RAC measures the \textit{sensitivity} of a model to a change in its shape: a model that perfectly segments the ground truth has $\mathrm{RAC} = 1$, a model that misses the change has $\mathrm{RAC} \approx 0$, and $\mathrm{RAC} > 1$ means the mask grows more than the added area.

\textbf{Comparison to human behavior.} Humans report a binary change/no-change decision, so the metric is a detection rate (DR). To compare with humans, we label a change as detected iff the relative change in mask area, $\delta = |A_{\mathrm{out}} - A_{\mathrm{init}}| / A_{\mathrm{init}}$, exceeds a threshold $\tau$, and report $\mathrm{DR}_{\mathrm{concave}}$ and $\mathrm{DR}_{\mathrm{convex}}$. We fit $\tau$ once on a reference model (SegFormer-B5 at its minimum-$\bar{E}$ checkpoint, selected on the TTC task, so no CD data are used for checkpoint selection), by minimizing the RMSE to the human detection rates over $\tau \in \{0.1, 1.1, \dots, 20.1\}\%$, which gives $\tau = 1.1\%$. We then apply it unchanged to every other configuration, which are therefore out of sample with respect to threshold selection. The convex--concave asymmetry is stable across the full range of $\tau$ (Appendix~\ref{app:detection_rate_vs_threshold}).

\section{Results}
\label{sec:results}\label{sec:human_like_representations_are_lim_networks}

Section~\ref{sec:human_like_representations} reports the TTC results: how alignment with humans changes over training and with model size, how the models compare with fitted geometric operators, and the mask geometry behind these effects. Section~\ref{sec:results_cd} tests the same geometry on change detection.

\subsection{Time-to-collision}
\label{sec:human_like_representations}

\begin{figure}[!htb]
  \centering
  \includegraphics[width=1\linewidth]{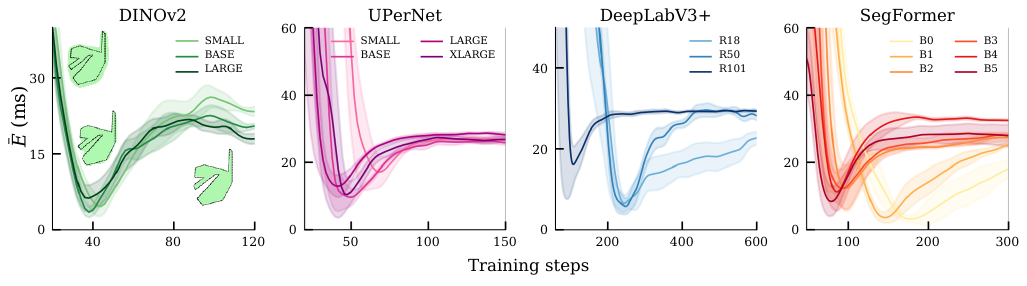}
  \caption{\textbf{Alignment with the human TTC effect peaks at intermediate training.} $\bar{E}$ (ms) across training, averaged over seeds; bands show $\pm$SEM, and darker lines are larger models. $\bar{E}$ stays above zero because the absolute value is taken per seed before averaging (Appendix~\ref{app:ttc_concavity_effect}).}
  \label{fig:ttc_overlays}
\end{figure}

\textbf{Alignment peaks at intermediate training.} In all four families, alignment with the human TTC effect is highest at an intermediate stage of training. $\bar{E}$ is high early in training, reaches a minimum mid-training, and rises again (Figure~\ref{fig:ttc_overlays}). Early masks fill concavities too much and give a concavity effect larger than the human one. Late masks resolve concavities and give a smaller one. Appendix~\ref{app:ttc_concavity_effect} plots this signed effect across training against the human value.

\textbf{The model effect passes through the human value.} On the way from coarse to fine, the model concavity effect passes through the human value in all 16 configurations, with the seed-averaged effect lying inside the human 95\% CI for a window of checkpoints. 

\textbf{The peak holds on held-out participants.} With the checkpoint selected on the other half of the participants, the best-aligned configurations reach $\bar{E}$ of 5--8\,ms. Testing all 16 configurations together, the selected checkpoint has lower $\bar{E}$ than both the first checkpoint ($t(15)=-5.17$, $p=5.7\times10^{-5}$) and the last one ($t(15)=-19.6$, $p=2.1\times10^{-12}$), so the trajectory is non-monotonic out of sample.

\textbf{Pretraining is not required.} DINOv2 trained from random initialization reaches the same held-out $\bar{E}$ as the pretrained backbone (5.5--6.3\,ms; Appendix~\ref{app:pretraining_ablation}).

\begin{figure}[!htb]
  \centering
  \includegraphics[width=1\linewidth]{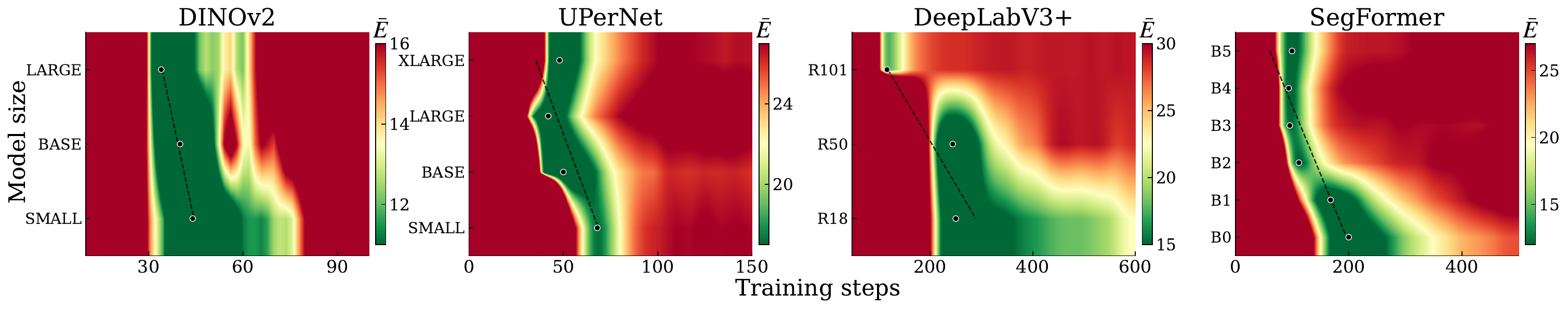}
  \caption{\textbf{Larger models reach low alignment error earlier.} $\bar{E}$ (ms) over training steps (horizontal) and model size (vertical) for each family. Low $\bar{E}$ forms a band that occurs earlier in training for larger models.}
  \label{fig:size_and_time_space}
\end{figure}

\textbf{Larger models reach the aligned regime earlier.}\label{sec:model_size_timing} In size--time space, low $\bar{E}$ forms a band that shifts to earlier training steps as model size grows (Figure~\ref{fig:size_and_time_space}). For example, SegFormer B0 reaches its minimum at step 202 and B5 at step 102, and DINOv2 Small at step 42 and Large at step 36 (Appendix Table~\ref{tab:ttc_best_checkpoints}). Per-family slopes are in Appendix~\ref{app:size_timing}.

\textbf{Fitted geometric operators describe the human effect equally well.}\label{sec:geometric_baselines} We applied classical operators to the ground-truth masks, fitted their single parameter to the human data, and scored them with the same TTC pipeline. Morphological closing reaches $\bar{E}=1.6$\,ms, $\alpha$-shapes 3.2\,ms, and dilation 4.8\,ms. These values are within the 0.8--14.5\,ms range of the models at their minimum-error checkpoints (Appendix~\ref{app:geometric_baselines}). The convex hull, which fills every concavity, overshoots by 412.8\,ms. This is consistent with a human body representation between the exact outline and the convex hull ~\citep{li2023approximate}. The two approaches differ in what they receive. The operators start from the correct segmentation, with a smoothing amount chosen on human responses. The models see only the image and no human data, and they pass through this geometry while learning to segment.

\begin{figure}[!htb]
  \centering
  \includegraphics[width=1\linewidth]{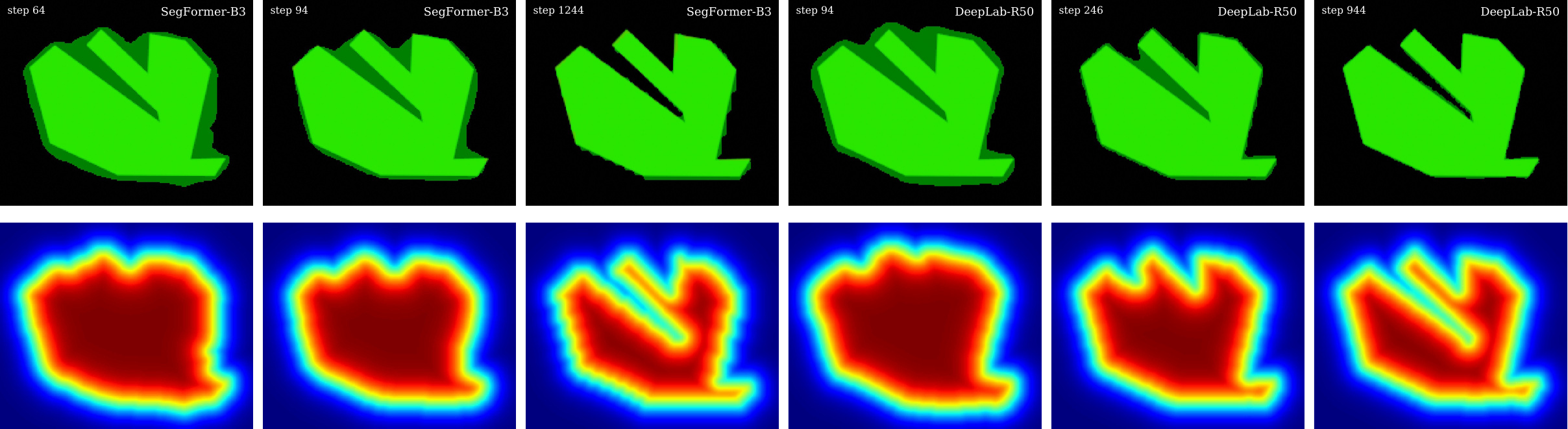}
  \caption{\textbf{Concavities are resolved after convex parts.} Masks (top) and foreground softmax probability maps (bottom) for SegFormer-B3 and DeepLabV3+-R50 at an early, an intermediate, and a late training step. Convex parts sharpen first; concave notches stay partly filled at intermediate steps and are resolved late.}
  \label{fig:train_collage}
\end{figure}

\textbf{Concavities are resolved last.}\label{sec:mask_geometry} At the aligned stage, convex parts are already sharp while concavities are only partly resolved. As training proceeds, convex corners become sharp first. Concavities stay partly filled for longer: the mask cuts across the mouth of each notch, and the probability map shows a wide low-confidence band inside it (Figure~\ref{fig:train_collage}). With more training, this band shrinks and the boundary moves into the notch; larger models show the same progression at a fixed step (Appendix~\ref{app:visual-representation}). Under translation, a partly filled concavity is reached earlier than a convex edge, which gives $\Delta>0$. The same geometry predicts that small additions inside a concavity change the mask less than additions on a convex part, which we test next.

\subsection{Change detection}
\label{sec:results_cd}

\begin{figure}[!htb]
    \centering
    \begin{subfigure}[b]{0.48\linewidth}
        \centering
        \includegraphics[width=\linewidth]{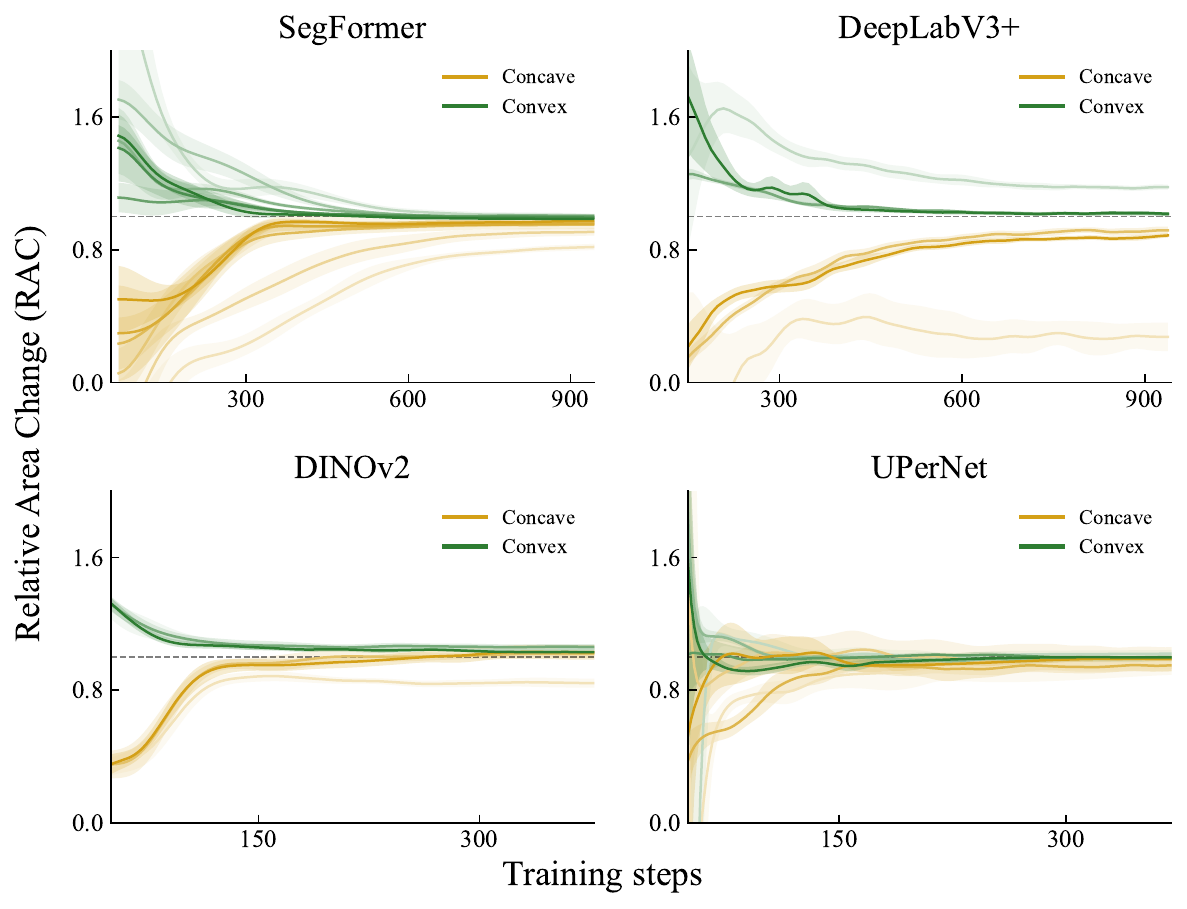}
        \caption{}
    \end{subfigure}
    \hfill
    \begin{subfigure}[b]{0.48\linewidth}
        \centering
        \includegraphics[width=\linewidth]{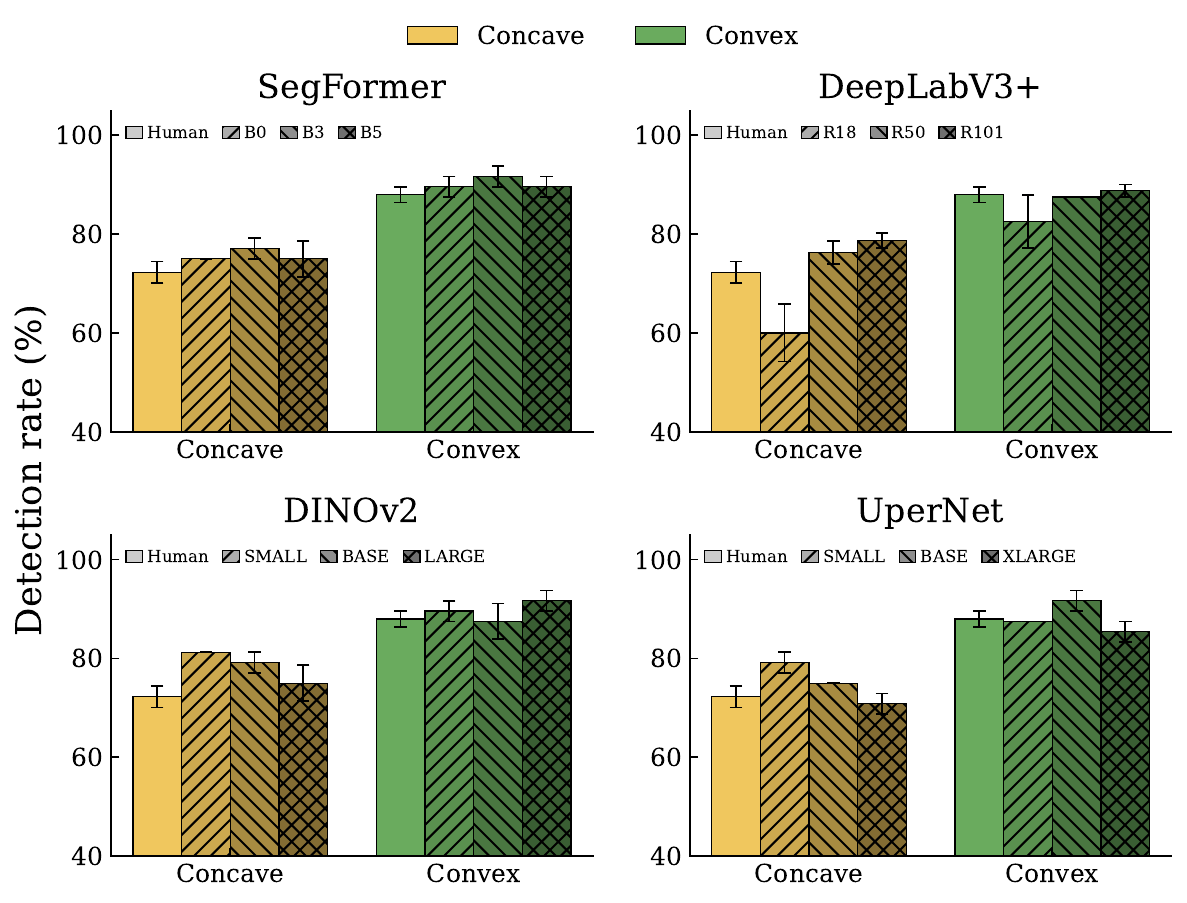}
        \caption{}
    \end{subfigure}
    \caption{\textbf{Models are less sensitive to concave than to convex changes, like humans.} \textbf{(A)} RAC across training; darker lines are larger models, and the dotted line marks RAC $=1$ (the added area is fully captured). Concave RAC is below convex RAC for most of training. \textbf{(B)} Detection rates with one threshold ($\tau = 1.1\%$) fitted on SegFormer-B5 and applied to all configurations shown.}
    \label{fig:rac_and_barplot}
\end{figure}

\textbf{Models are less sensitive to concave changes.} Like humans, the models are less sensitive to changes inside concavities than to changes on convex parts. Concave RAC stays below convex RAC for most of training and approaches it as concavities are resolved, earlier for larger models (Figure~\ref{fig:rac_and_barplot}A).

\textbf{Detection rates follow the human ordering.} Converted to binary decisions, the models detect convex changes more often than concave ones in all four families, as humans do (Figure~\ref{fig:rac_and_barplot}B). The detection threshold $\tau=1.1\%$ is fitted once, on SegFormer-B5 at its TTC-selected checkpoint, and applied unchanged to the other 11 configurations. The ordering holds over a wide range of $\tau$ (Appendix~\ref{app:cd_analyses}).

\FloatBarrier
\section{Discussion}
\label{sec:discussion}
People often rely on approximate ``body'' representations when making physical judgments: coarse, volumetric approximations that discard fine detail and texture. Fitted geometric operators reproduce the resulting concavity effect (Section~\ref{sec:geometric_baselines}), but they do not explain how such geometry could be learned. We therefore asked whether similar geometry arises in deep segmentation models during standard training.

We find that it does. Model masks simplify concavities while preserving convex detail, mirroring the asymmetry in human responses. Early in training, masks smooth over concavities; with continued training, they resolve increasingly precise boundaries. Alignment with humans peaks in between: mid-training models attain the lowest alignment error, and larger models reach this regime earlier, a trend consistent across families. Some architectures and training regimes produce effects resembling human judgments, and we illustrate candidate geometries consistent with those effects. Object approximations consistent with human concavity effects thus emerge from segmentation training alone, without any human behavioral supervision, and segmentation accuracy and human alignment peak at different points along a coarse-to-fine trajectory.

Why would humans rely on such approximations? The human visual system develops under metabolic and informational constraints, time pressure, and limited perceptual resolution, which may favor efficient, lower-complexity representations. Our results fit this resource-rational view ~\citep{lieder2020resource}: with limited training, models also favor coarse geometry over fine detail, and this is the stage at which their behavior best matches human judgments. Segmentation models therefore offer a concrete, controllable setting for studying how such representations can arise.

Precise segmentation remains essential for tasks that depend on exact contours, such as instance annotation or fine-grained recognition. When a model's judgments should agree with people's, however, intermediate coarseness gives the closest match, consistent with the coarse-body account of \citet{li2023approximate}. Checkpoint selection is therefore a simple way to obtain masks that better match human physical judgments from standard segmentation models, with potential value for future work in applications sensitive to agreement with human judgment, such as domestic human--robot interaction.

\section{Limitations and Conclusion}
Our experiments use 2D polygon stimuli and controlled kinematics: models segment a single static frame and receive ground-truth velocities. This choice is deliberate, because it isolates object representations from motion estimation and learned dynamics. Humans are also sensitive to the ground-truth time to collision: their concavity effect grows from 20.7 to 53.0 ms as collision time increases. Our models process one static frame and cannot adapt to this dependence, so we aggregate across collision times. Finally, fitted geometric operators applied to the ground-truth masks match the models on the alignment metric (Section~\ref{sec:geometric_baselines}). The models should therefore not be read as a better fit to the human summary; their value lies in showing how such representations could arise in a general-purpose perceptual system.
Future work should test these effects in realistic 3D environments and in more advanced model paradigms, such as vision-language models.

\textbf{Conclusion.} Behavioral evidence suggests that people rely on coarse object bodies, not precise shapes, when predicting everyday physical events. We show that segmentation models pass through a regime whose behavioral consequences resemble these human effects, at intermediate stages of training and earlier in larger models. Theoretically, this suggests that a general visual system, not a hand-engineered geometry, can develop human-like object representations when training or size is limited, consistent with resource-rational accounts of cognition ~\citep{lieder2020resource}. Practically, we provide a framework for measuring how closely model object representations align with human ones. The framework shows that higher segmentation accuracy does not imply closer alignment with human judgments.

\clearpage

\section*{Reproducibility Statement}
Code and the synthetic fine-tuning dataset are available in an anonymized repository at \url{https://anonymous.4open.science/r/shape-alignment-3C2D}. The repository includes the procedural polygon generator (Appendix~\ref{app:synthetic-dataset}), the fine-tuning configuration for all four model families (Appendix~\ref{app:finetuning_hyperparameters}), the TTC and CD evaluation pipeline (Sections~\ref{sec:ttc_pipeline} and \ref{sec:cd_pipeline}), and the analysis scripts for the held-out checkpoint selection, the non-monotonicity tests, and the geometric baselines (Sections~\ref{sec:human_like_representations} and \ref{sec:results_cd}). The human behavioral data are the publicly released data of \citet{li2023approximate}; the participant-level re-analysis that yields $\Delta_{\mathrm{human}}$ and its confidence interval is included in the repository.

\section*{AI Use Statement}
We used generative AI tools (large language models) for editing and rewriting text; checking consistency of numbers across sections; drafting LaTeX for tables; and drafting and debugging analysis code. All experiments, analyses and results were produced by the authors, who verified every AI-assisted change and take full responsibility for the content.

\newpage
{
    \small
    \bibliographystyle{iclr2027_conference}
    \bibliography{iclr2027_conference}
}

\newpage

\appendix
\section*{Appendix overview}

\begin{itemize}
    \item \textbf{Appendix~\ref{app:human_data}: Human data.} The two experiments of \citet{li2023approximate}, the CD stimuli, and the summary statistics of the human effect.
    \item \textbf{Appendix~\ref{app:models_training}: Models and training.} Architectures, the synthetic fine-tuning dataset, and the training settings.
    \item \textbf{Appendix~\ref{app:visual-representation}: Mask geometry across model size.} How masks change with model size at a fixed training step.
    \item \textbf{Appendix~\ref{app:ttc_analyses}: Time-to-collision analyses.} Statistics at the minimum-error checkpoints, held-out checkpoint selection with per-family tests, the signed effect across training, the size--timing slopes, and the geometric baselines.
    \item \textbf{Appendix~\ref{app:pretraining_ablation}: Pretraining ablation.} DINOv2 trained from random initialization.
    \item \textbf{Appendix~\ref{app:cd_analyses}: Change-detection analyses.} Threshold selection and robustness to the threshold.
\end{itemize}

\section{Human data}
\label{app:human_data}

All human data are the publicly released data of \citet{li2023approximate}. The human stimuli are used only for evaluation; no model is trained on them.

\textbf{Time-to-collision (TTC).} 178 participants watched 96 videos in which one polygon moves toward a stationary one. At a known time the stationary polygon disappears, and participants press a key when they think the two have collided. The response time is measured from motion onset to the key press. The videos form 48 matched pairs (8 shapes $\times$ 6 ground-truth collision times). The two videos of a pair share the same kinematics and differ only in whether a concave or a convex side of the stationary polygon faces the collision.

\textbf{Human TTC effect.} Each participant contributes one convex$-$concave difference. Averaged over participants, the concavity effect is $\Delta_{\mathrm{human}} = 31.17$\,ms (95\% CI $[22.74, 39.60]$, $t$-interval over the 178 per-participant differences). The between-participant standard deviation is 57.0\,ms, so the effect corresponds to Cohen's $d = 0.55$. The effect depends on the ground-truth collision time: it grows from 20.7 to 53.0\,ms as collision time increases. Our models process a single static frame and cannot express this dependence, so all model--human comparisons average over collision times (Equation~\ref{eq:ttc_signed_error}).

\begin{figure}[!htbp]
    \centering
    \includegraphics[width=0.3\linewidth]{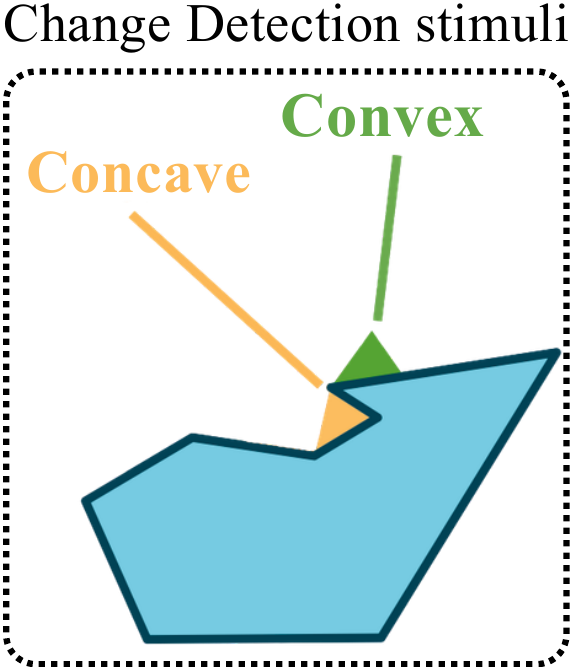}
    \caption{\textbf{Change-detection stimuli.} The same area is added either inside a concavity (orange) or on a convex part (green).}
    \label{fig:cd_stimuli}
\end{figure}

\textbf{Change detection (CD).}\label{app:cd_stimuli} 50 participants saw a polygon $I_{\mathrm{init}}$ for 1\,s, a blank screen for 2\,s, and a second polygon $I_{\mathrm{out}}$ for 1\,s, and reported whether it had changed. $I_{\mathrm{out}}$ adds a small triangle of equal area either inside a concavity or on a convex part of $I_{\mathrm{init}}$ (Figure~\ref{fig:cd_stimuli}), with 16 stimuli per condition (8 shapes $\times$ 2 change sizes). \citet{li2023approximate} also include a third condition, in which the added area lies inside a large concavity without filling it; we use only the concave and convex conditions. Humans detect convex changes more often than concave ones. This is Experiment~3b of \citet{li2023approximate}.

\FloatBarrier

\section{Models and training}
\label{app:models_training}

\subsection{Architectures}
\label{app:architectures}

We evaluate 16 configurations from four families (Table~\ref{tab:architectures}). The families cover convolutional backbones (DeepLabV3+ with ResNet, UPerNet with ConvNeXt) and transformer backbones (SegFormer, DINOv2), each at several sizes. All models are fine-tuned end to end on the synthetic dataset of Appendix~\ref{app:synthetic-dataset} with the same recipe (Appendix~\ref{app:finetuning_hyperparameters}). Per-configuration parameter counts are listed in Table~\ref{tab:ttc_best_checkpoints}.

\begin{table}[!htbp]
    \centering
    \small
    \setlength{\tabcolsep}{4pt}
    \begin{tabular}{llcl}
    \toprule
    Family & Backbones & Params (M) & Initialization \\
    \midrule
    SegFormer~\citep{xie2021segformer} & B0--B5 & 3.7--84.6 & ADE20K~\citep{zhou2017scene} \\
    DeepLabV3+~\citep{chen2018encoder} & ResNet-18/50/101 & 12.3--60.4 & Cityscapes~\citep{cordts2016cityscapes} \\
    UPerNet~\citep{xiao2018unified} & ConvNeXt S/B/L/XL & 82.0--391.0 & ADE20K \\
    DINOv2$^{\dagger}$~\citep{oquab2024dinov2} & S/B/L & 27.2--310.6 & Self-supervised \\
    \bottomrule
    \end{tabular}
    \caption{\textbf{Model families.} Parameter ranges span the smallest to the largest backbone in each family. $^{\dagger}$With a light convolutional decoder.}
    \label{tab:architectures}
\end{table}

\subsection{Synthetic fine-tuning dataset}
\label{app:synthetic-dataset}

Models pretrained on natural images do not reliably segment the polygonal stimuli of \citet{li2023approximate}. To adapt them without contaminating evaluation, we built a synthetic dataset that matches the geometric and photometric statistics of the human stimuli but shares no shapes with them. The dataset and its generator are in the anonymized repository linked in the Reproducibility Statement.

\textbf{Canvas.} Each image is a $1000 \times 400$ (width $\times$ height) RGB canvas with a single uniformly colored polygon centered on a black background. The aspect ratio matches the stimuli of \citet{li2023approximate}.

\textbf{Polygon generator.} Each polygon is defined by:
\begin{itemize}
    \item a vertex count sampled uniformly from $\{5, \dots, 12\}$;
    \item a base radius $r_{\mathrm{base}} \sim \mathcal{U}(58, 62)$ pixels, which keeps polygon area in a narrow band;
    \item an irregularity coefficient $\sim \mathcal{U}(0.15, 0.30)$ for the angular jitter between successive vertices;
    \item a spikiness coefficient $\sim \mathcal{U}(0.15, 0.25)$, which sets the radial variation around $r_{\mathrm{base}}$.
\end{itemize}
After this polar layout, each vertex is moved by an independent Cartesian offset of magnitude $\sim \mathcal{U}(0.20, 0.40)\cdot r_{\mathrm{base}}$, so that shapes are not star-convex. Independent $x$ and $y$ scale factors $\sim \mathcal{U}(0.85, 1.15)$ then remove any remaining circular bias.

\textbf{Concavities.} Each polygon receives 1 to 4 concave notches (uniform). A notch is inserted on a random edge by adding interior vertices pulled 35--55\% of the way toward the centroid, with small lateral jitter. Three notch styles are mixed:
\begin{itemize}
    \item one interior vertex near the edge midpoint (50\% of notches);
    \item two interior vertices at $1/3$ and $2/3$ of the edge (30\%);
    \item two interior vertices close together around the midpoint (20\%).
\end{itemize}
A polygon is rejected and resampled if it leaves the canvas (10\,px margin) or if any two non-adjacent edges intersect. Together with the offsets and scaling above, this covers the concave geometry of the human stimuli without reproducing any of them.

\textbf{Colors.} Fill colors are drawn uniformly from 24 saturated hues, luminance-matched to the stimuli of \citet{li2023approximate}: RGB channels are capped near $V \approx 81\%$, saturation is near 100\%, and per-color luminance falls in high (140--191), medium (85--139), and low (17--64) bands in a 40/40/20 ratio. Mean luminance is $\approx 112$, close to that of the human stimuli. Polygons are rendered at $4\times$ supersampling and downsampled with Lanczos resampling, giving antialiased edges.

\textbf{Splits and ground truth.} The dataset has 500 training and 200 validation images, split with a fixed seed. Shapes and colors are sampled independently per image, so no polygon is shared across splits, and no frame of \citet{li2023approximate} appears in either split. Ground-truth masks are obtained by thresholding the rendered image at intensity $>10$, which separates the polygon from the background while keeping antialiased boundary pixels. No data augmentation is used; training and validation losses agree closely throughout fine-tuning.

\subsection{Fine-tuning hyperparameters}
\label{app:finetuning_hyperparameters}

All configurations share the settings in Table~\ref{tab:finetuning_hyperparameters}.

\begin{table}[!htbp]
    \centering
    \small
    \begin{tabular}{ll}
    \toprule
    Setting & Value \\
    \midrule
    Optimizer & AdamW (weight decay 0.01) \\
    Learning rate & $5 \times 10^{-5}$ \\
    Scheduler & Cosine annealing with linear warmup (5\% of total steps) \\
    Loss & Cross-entropy + Dice ($\lambda = 0.5$) \\
    Effective batch size & 8 (gradient accumulation) \\
    Training length & 15 epochs (945 steps) \\
    Training / validation images & 500 / 200 \\
    Data augmentation & None \\
    Image size & $400 \times 1000$ px (height $\times$ width) \\
    Precision & bfloat16 (mixed) \\
    Gradient clipping & Max norm 1.0 \\
    Seeds & 3 per configuration \\
    Hardware & NVIDIA A100 (80\,GB) \\
    \bottomrule
    \end{tabular}
    \caption{\textbf{Fine-tuning hyperparameters,} shared by all 16 configurations.}
    \label{tab:finetuning_hyperparameters}
\end{table}

\FloatBarrier

\section{Mask geometry across model size}
\label{app:visual-representation}

\begin{figure}[!htbp]
    \centering
    \includegraphics[width=0.92\textwidth]{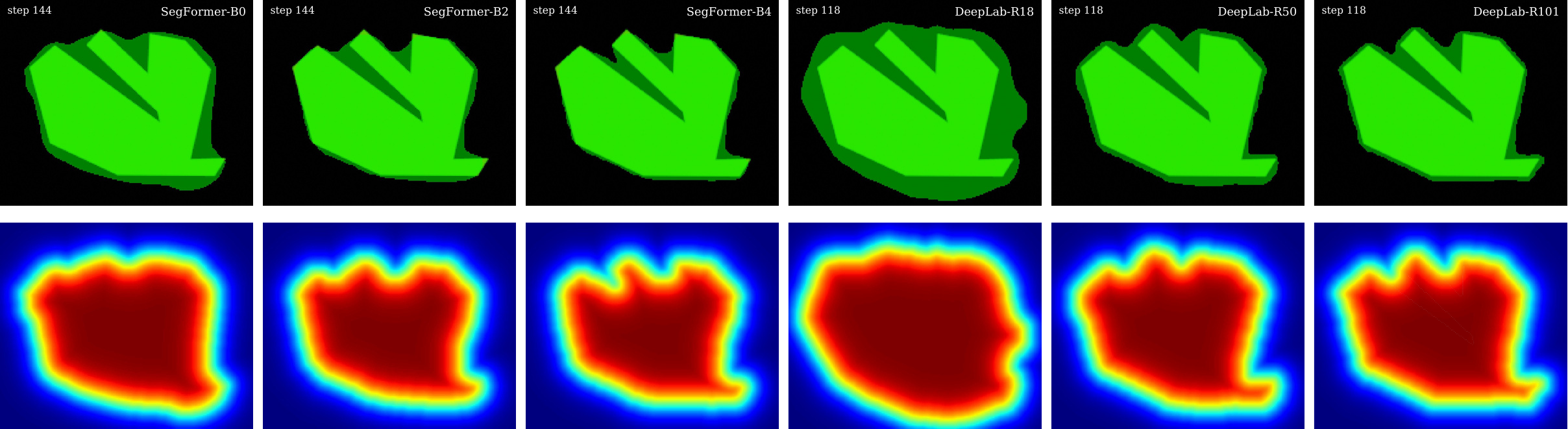}
    \caption{\textbf{Masks across model size,} at a fixed training step (step 144 for SegFormer B0/B2/B4; step 118 for DeepLabV3+ R18/R50/R101). Top: mask overlays. Bottom: foreground probability maps, obtained applying a softmax to the logits.}
    \label{fig:collage_size}
\end{figure}

Figure~\ref{fig:collage_size} shows masks at a fixed training step for three sizes in two families. Smaller models produce coarser masks that leave concave notches partly filled. Larger models resolve the notches further at the same step, while convex parts are sharp at all sizes. This is the same progression that training time produces (Figure~\ref{fig:train_collage}).

\FloatBarrier

\section{Time-to-collision analyses}
\label{app:ttc_analyses}

\subsection{Statistics at minimum-error checkpoints}
\label{app:ttc_best_checkpoint_table}

Table~\ref{tab:ttc_best_checkpoints} describes each configuration at the checkpoint with minimum $\bar{E}$ on the full participant sample. These numbers are in-sample; held-out results are in Appendix~\ref{app:heldout}. $\bar{E}$ and $\Delta$ are given as mean $\pm$ SEM across seeds, with $\Delta$ first averaged over the six collision times. \emph{Pairwise} is the number of the 48 matched concave--convex pairs (8 shapes $\times$ 6 collision times) with $\mathrm{TTC}_{\mathrm{convex}} > \mathrm{TTC}_{\mathrm{concave}}$, averaged across seeds. \emph{Test} is a one-sided paired $t$-test over matched pairs, $H_1\!: \mu_{\mathrm{convex}} > \mu_{\mathrm{concave}}$ ($^{*}p<.05$, $^{**}p<.01$, $^{***}p<.001$).

Every configuration shows a positive concavity effect at its minimum-error checkpoint. Over the whole trajectory, the seed-averaged effect lies inside the human 95\% CI for a window of checkpoints in all 16 configurations (Table~\ref{tab:trajectory_stats}).

\begin{table}[!htbp]
    \centering
    \small
    \begin{tabular}{lrrccrc}
    \toprule
    Model & Params (M) & Best step & $\bar{E}$ (ms) & $\Delta$ (ms) & Pairwise & Test \\
    \midrule
    SegFormer B0 & 3.7 & 202 & $2.4 \pm 0.5$ & $31.3 \pm 1.8$ & 45/48 & *** \\
    SegFormer B1 & 13.7 & 164 & $5.3 \pm 1.4$ & $32.4 \pm 3.3$ & 43/48 & *** \\
    SegFormer B2 & 27.4 & 102 & $5.0 \pm 3.4$ & $28.6 \pm 4.3$ & 44/48 & *** \\
    SegFormer B3 & 47.2 & 98 & $4.9 \pm 2.7$ & $27.0 \pm 3.1$ & 46/48 & *** \\
    SegFormer B4 & 64.1 & 92 & $3.4 \pm 1.3$ & $31.0 \pm 2.3$ & 46/48 & *** \\
    SegFormer B5 & 84.6 & 102 & $3.0 \pm 1.8$ & $32.1 \pm 2.5$ & 46/48 & *** \\
    \midrule
     DeepLabV3+ R18 & 12.3 & 240 & $3.8 \pm 1.6$ & $34.3 \pm 2.2$ & 39/48 & *** \\
    DeepLabV3+ R50 & 41.3 & 248 & $2.5 \pm 1.1$ & $29.3 \pm 1.7$ & 31/48 & ** \\
    DeepLabV3+ R101 & 60.4 & 94 & $6.0 \pm 5.3$ & $26.0 \pm 5.8$ & 35/48 & ** \\
    \midrule
    DINOv2 Small & 27.2 & 42 & $3.2 \pm 1.9$ & $29.6 \pm 2.8$ & 44/48 & *** \\
    \textbf{DINOv2 Base} & \textbf{92.9} & \textbf{38} & $\mathbf{0.8 \pm 0.3}$ & $\mathbf{31.2 \pm 0.6}$ & \textbf{46/48} & *** \\
    DINOv2 Large & 310.6 & 36 & $3.1 \pm 2.3$ & $28.7 \pm 2.7$ & 47/48 & *** \\
    \midrule
    UPerNet Small & 82.0 & 68 & $14.5 \pm 5.9$ & $18.5 \pm 7.8$ & 38/48 & *** \\
    UPerNet Base & 122.0 & 38 & $3.3 \pm 1.8$ & $34.4 \pm 1.8$ & 41/48 & *** \\
    UPerNet Large & 235.0 & 40 & $11.1 \pm 4.0$ & $20.0 \pm 4.0$ & 42/48 & *** \\
    UPerNet XLarge & 391.0 & 42 & $4.3 \pm 2.7$ & $26.8 \pm 2.7$ & 42/48 & *** \\
    \midrule
    Human & & & & $31.17$ [22.74, 39.60] & & \\
    \bottomrule
    \end{tabular}
    \caption{\textbf{TTC statistics at minimum-error checkpoints (in-sample).} The best configuration is in bold. The human row gives the mean effect and its 95\% CI.}
    \label{tab:ttc_best_checkpoints}
\end{table}

\subsection{Held-out checkpoint selection}
\label{app:heldout}

\textbf{Protocol.} We split the 178 participants into two equal folds. The minimum-$\bar{E}$ checkpoint is selected on one fold, with $\Delta_{\mathrm{human}}$ computed from that fold only, and $\bar{E}$ at that checkpoint is measured on the other fold. This is repeated 200 times, in both directions.

\textbf{Results.} On held-out participants, the best-aligned configuration of each family keeps $\bar{E}$ of 5--8\,ms (SegFormer B0 5.3, UPerNet XLarge 5.6, DINOv2 Base 5.6, DeepLabV3+ R50 7.6\,ms). Selecting on held-out participants instead of on the evaluated ones raises $\bar{E}$ by 0.4--5.4\,ms across the 16 configurations, less than the 8.4\,ms half-width of the human 95\% CI.

\textbf{Test of non-monotonicity.} On each held-out fold, we compare $\bar{E}$ at the selected checkpoint with $\bar{E}$ at the first and last checkpoints, which are fixed in advance. A non-monotonic trajectory requires the selected checkpoint to be lower than both. The main text reports the test across all 16 configurations. Table~\ref{tab:trajectory_stats} also gives the tests within each family. With three to six configurations per family, these tests have low power. The comparison with the last checkpoint is significant in all four families, and the comparison with the first checkpoint in SegFormer, UPerNet and DeepLabV3+.

\begin{table}[!htbp]
  \centering
  \caption{\textbf{Per-family tests of the training trajectory.} The aligned checkpoint is selected on one half of the participants, and $\bar{E}$ is compared on the other half with the first and last checkpoints (one-sided paired $t$-tests across the $n$ backbones of each family, $df = n-1$). $\Delta$ in human CI: configurations whose seed-averaged concavity effect lies inside the human 95\% CI for a window of checkpoints. $\hat{\beta}$: OLS slope of model-size rank on the minimum-$\bar{E}$ training step, in size ranks per step; negative values mean larger models reach the minimum earlier (two-sided $p$, $df = n-2$; Appendix~\ref{app:size_timing}).}
  \label{tab:trajectory_stats}
  \small
  \setlength{\tabcolsep}{5pt}
  \begin{tabular}{lc rr rr c rr}
    \toprule
    & & \multicolumn{2}{c}{Aligned $<$ first} & \multicolumn{2}{c}{Aligned $<$ last} & & \multicolumn{2}{c}{Size slope} \\
    \cmidrule(lr){3-4} \cmidrule(lr){5-6} \cmidrule(lr){8-9}
    Family & $n$ & $t$ & $p$ & $t$ & $p$ & $\Delta$ in human CI & $\hat{\beta}$ & $p$ \\
    \midrule
    SegFormer  & 6  & $-2.68$ & $.022$ & $-27.4$ & ${<}.0001$ & 6/6 & $-0.037$ & $.020$ \\
    UPerNet    & 4  & $-4.15$ & $.013$ & $-8.03$ & $.002$     & 4/4 & $-0.090$ & $.216$ \\
    DINOv2     & 3  & $-2.54$ & $.063$ & $-31.6$ & ${<}.001$  & 3/3 & $-0.197$ & $.073$ \\
DeepLabV3+ & 3  & $-5.25$ & $.017$ & $-12.6$ & $.003$     & 3/3 & $-0.012$ & $.308$ \\
    \midrule
     All        & 16 & $-5.17$ & $5.7\times10^{-5}$ & $-19.6$ & $2.1\times10^{-12}$ & 16/16 & -- & -- \\
    \bottomrule
  \end{tabular}
\end{table}

\textbf{Seed variability.} At the selected checkpoint, the standard deviation of $\bar{E}$ across training seeds is 4.3\,ms on average, below the 8.4\,ms half-width of the human CI. Per-configuration seed spread is the SEM column of Table~\ref{tab:ttc_best_checkpoints}.

\textbf{Effect sizes.} In units of the human between-participant standard deviation (57.0\,ms), the human effect is $d=0.55$, and the model effects at their selected checkpoints span $d=0.32$ to $0.60$.

\subsection{Signed concavity effect across training}
\label{app:ttc_concavity_effect}

\begin{figure}[!htbp]
    \centering
    \includegraphics[width=\linewidth]{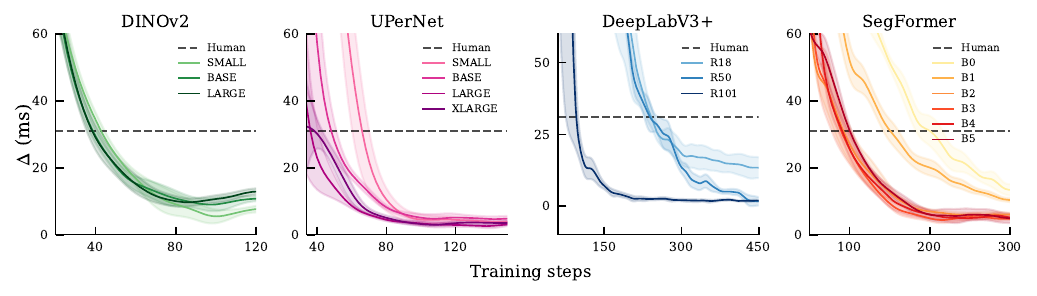}
    \caption{\textbf{Signed TTC concavity effect across training.} Positive values mean earlier predicted collision for concave than for convex configurations. Lines: mean across seeds; bands: $\pm$SEM; dashed line: human effect.}
    \label{fig:ttc_concavity_effect_appendix}
\end{figure}

Figure~\ref{fig:ttc_concavity_effect_appendix} shows the signed effect $\Delta_{\mathrm{model},s}(\tau)$, averaged over collision times and then over seeds. Across families, $\Delta$ starts above the human value, passes through it, and then falls further as concave boundaries are resolved. In all 16 configurations, the seed-averaged effect lies inside the human 95\% CI for a window of checkpoints.

\textbf{Why $\bar{E}$ does not reach zero.} The signed error of each seed does pass through zero. Without smoothing, the minimum of $|E_s|$ over checkpoints is 0.17--1.03\,ms across the SegFormer backbones, and individual seeds reach 0.01\,ms. The plotted curves stay higher for two reasons:
\begin{itemize}
    \item $\bar{E}$ takes the absolute value per seed and then averages. Seeds cross zero at different steps, so the average does not touch zero even when each seed does. Averaging the signed errors first would give 0.04--1.54\,ms.
    \item Display smoothing adds a further 1.1--7.0\,ms.
\end{itemize}
Seed averaging is the larger of the two. In addition, checkpoints are sampled at discrete steps, so none needs to fall exactly at a zero crossing.

\subsection{Model size and timing of alignment}
\label{app:size_timing}

For each family, $\hat{\beta}$ is the OLS slope of model-size rank on the
minimum-$\bar{E}$ training step (Table~\ref{tab:trajectory_stats}).

The fitted steps are the per-row minima of the Figure~\ref{fig:size_and_time_space} heatmaps (seed-averaged $\bar{E}$, smoothed), not the unsmoothed best steps of Table~\ref{tab:ttc_best_checkpoints}.
It is negative in all four families and significant for SegFormer
($n=6$, $p=.020$), but not for UPerNet ($p=.216$), DINOv2 ($p=.073$) or
DeepLabV3+ ($p=.308$). With three to six sizes per family, these
per-family tests have little power.

\subsection{Geometric baselines}
\label{app:geometric_baselines}

We applied four operators to the ground-truth polygon masks of the TTC stimuli: morphological closing, $\alpha$-shapes ~\citep{edelsbrunner1983shape}, dilation, and the convex hull. The free parameter of each operator was fitted to the human data, and the resulting masks were scored with the same TTC pipeline and $\bar{E}$ as the models (Table~\ref{tab:geometric_baselines}). Both the operator parameters and the model checkpoints are selected on the full participant sample, so the comparison is in-sample on both sides.

\begin{table}[!htbp]
    \centering
    \small
    \begin{tabular}{lcr}
    \toprule
    Operator & Fitted parameter & $\bar{E}$ (ms) \\
    \midrule
    Morphological closing & $r = 36$ & 1.6 \\
    $\alpha$-shape & $\alpha = 24$ & 3.2 \\
    Dilation & $r = 6$ & 4.8 \\
    Convex hull & none & 412.8 \\
    \midrule
    Models, minimum-error checkpoints (Table~\ref{tab:ttc_best_checkpoints}) & & 0.8--14.5 \\
    \bottomrule
    \end{tabular}
    \caption{\textbf{Geometric operators applied to ground-truth masks,} with parameters fitted to the human data and scored with the same $\bar{E}$ as the models.}
    \label{tab:geometric_baselines}
\end{table}

\FloatBarrier

\section{Pretraining ablation}
\label{app:pretraining_ablation}

\textbf{Motivation.} All models start from weights learned on natural images: SegFormer, UPerNet, and DeepLabV3+ from semantic-segmentation checkpoints (ADE20K or Cityscapes), and DINOv2 from self-supervised weights (Table~\ref{tab:architectures}). The coarse-to-fine trajectory could therefore be inherited from this pretraining rather than produced by learning to segment the polygons.

\textbf{Setup.} We reran the full DINOv2 pipeline from random initialization: fine-tuning on the synthetic dataset (Appendix~\ref{app:synthetic-dataset}) and TTC evaluation on the human stimuli. Both the pretrained and the randomly initialized backbones were scored with the held-out participant protocol of Appendix~\ref{app:heldout}.

\textbf{Result.} Held-out $\bar{E}$ is 5.5--6.3\,ms for both the pretrained and the randomly initialized backbones. Natural-image pretraining is therefore not needed to reach the level of alignment reported in the main text.

\FloatBarrier

\section{Change-detection analyses}
\label{app:cd_analyses}

\subsection{From mask areas to detection decisions}
\label{app:cd_threshold}

Section~\ref{sec:cd_pipeline} uses two area measures. RAC $= (A_{\mathrm{out}} - A_{\mathrm{init}})/A_{\mathrm{gt}}$ normalizes by the area of the added segment and measures local sensitivity (main-text Figure~\ref{fig:rac_and_barplot}A). To obtain binary decisions comparable to human responses, we instead use the relative change in mask area, $\delta = |A_{\mathrm{out}} - A_{\mathrm{init}}|/A_{\mathrm{init}}$, and label a change as detected iff $\delta > \tau$.

\textbf{Threshold selection.} The threshold is fitted once, on a reference configuration fixed in advance: SegFormer-B5 (ADE20K, seed 123) at its minimum-$\bar{E}$ checkpoint. That checkpoint is selected on the TTC task, so no CD data are used for checkpoint selection. We choose $\tau$ by minimizing the RMSE between model and human detection rates over $\tau \in \{0.1, 1.1, \dots, 20.1\}\%$, which gives $\tau = 1.1\%$. This value is then applied unchanged to every other configuration (main-text Figure~\ref{fig:rac_and_barplot}B), which are therefore out of sample with respect to the threshold.

\subsection{Detection rate vs.\ threshold}
\label{app:detection_rate_vs_threshold}

\begin{figure}[H]
    \centering
    \includegraphics[width=0.9\textwidth]{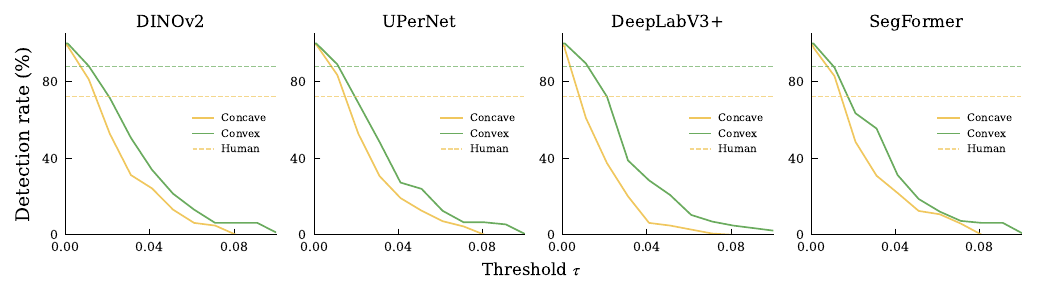}
    \caption{\textbf{Detection rate as a function of the threshold $\tau$ on $\delta$}, at the final training step, averaged over backbones and seeds. Solid lines: model detection rates for concave (yellow) and convex (green) changes. Dashed lines: human detection rates.}
    \label{fig:dr_vs_threshold}
\end{figure}

The convex--concave ordering does not depend on the chosen threshold. In all four families, convex changes are detected more often than concave ones over a wide range of $\theta$ (Figure~\ref{fig:dr_vs_threshold}). This figure uses the final training step, not the minimum-$\bar{E}$ checkpoint.

\end{document}